# A Hybrid Bacterial Foraging Algorithm For Solving Job Shop Scheduling Problems

**Narendhar. S and Amudha. T**

## Abstract

*Bio-Inspired computing is the subset of Nature-Inspired computing. Job Shop Scheduling Problem is categorized under popular scheduling problems. In this research work, Bacterial Foraging Optimization was hybridized with Ant Colony Optimization and a new technique Hybrid Bacterial Foraging Optimization for solving Job Shop Scheduling Problem was proposed. The optimal solutions obtained by proposed Hybrid Bacterial Foraging Optimization algorithms are much better when compared with the solutions obtained by Bacterial Foraging Optimization algorithm for well-known test problems of different sizes. From the implementation of this research work, it could be observed that the proposed Hybrid Bacterial Foraging Optimization was effective than Bacterial Foraging Optimization algorithm in solving Job Shop Scheduling Problems. Hybrid Bacterial Foraging Optimization is used to implement real world Job Shop Scheduling Problems.*

## Keywords:

## 1. INTRODUCTION

Optimization is a mathematical order, which involves the operation of finding minimum and maximum of functions. Optimization originated in 1990's, when George Danzig used Mathematical techniques for generating programs for Military applications. A calculation problem in which, the object is to find the best of all achievable solutions [39]. More formally, find a solution in the feasible region which has the min or (max) value of the objective function.

Optimization Problems includes Combinatorial Optimization, Heuristics, Metaheuristics, NP-Hard Problem [34], NP-Complete Problem. Approximation, Randomization, Restriction, Parameterization are the few methods to solve Optimization Problems. Scheduling is a jobs, machines and processing times. The schedule is subject to feasibility constraints and optimization objectives. Scheduling constraints are (i) Each machine can only process one job at a time. (ii) At any time, each job can only be processed by one machine. (iii) After completing the current job, the machine will move to the next job. Nature-Inspired computing paves way to develop new computing technique which is based on nature behaviour in solving complex problems.

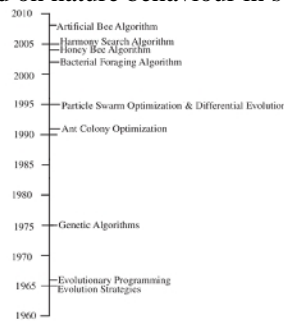

Figure 1: Popular nature-inspired meta-heuristics in chronological order

### 1.1 Job Shop Scheduling Problem (JSSP)

| Job | : | A piece of work that goes through series of operations. |
| Shop | : | A place for manufacturing or repairing of goods or machinery. |
| Scheduling | : | Decision process aiming to deduce the order of processing. |

The JSSP is an operation sequencing problem on multiple machine subject to some precedence constraints among the operations. The JSSP can be described as a set of n jobs denoted by $J_j$ where j =1,2…n which have to be processed on a set of m machines denoted by $M_k$ where k =1,2….m. Operation of $j^{th}$ job on the $k^{th}$ machine will be denoted by $O_{jk}$ with the processing time $p_{jk}$ [19] .Each job should be processed through the machines in a particular order or also known as technological constraint. Once a machine starts to process a job, processing of an operation cannot be interrupted. The required time to complete all operations for their processes is called makespan. JSSP are widely known as NP-Hard problem.

### Constraints

The JSSP subjects to two constraints, known as the operation precedence constraint and machine processing constraint:

1) The operation precedence constraint: The job is fixed in the order of operations and the processing of an operation cannot be interrupted and parallel.
2) The machine processing constraint: Only a single job can be processed at the identical time on the identical machine.

The main factor affecting to the JSSP is the nature of job shop. In static and deterministic job shop, all jobs are accessible for processing without delay, and no new jobs appear over time. In the dynamic probabilistic job shop, jobs appear arbitrarily over time, and processing times are probabilistic. This is more reasonable job shop situation but more difficult to solve it.

## 1.2 Ant Colony Optimization

ACO algorithm first proposed by M. Dorigo, in 1992 [29]. It is a metaheuristic in which a colony of ants capable of finding shortest path from their nest to food sources using pheromone trials. Ant probably coming later chooses the path is proportional to the quantity of pheromone on the path, earlier deposited by other ants.

Ants want to find food, so they set off from their nest and arrive at a decision point at which they have to decide which path to go on, for there are three different paths. Since they have no clue about which is the best choice, they choose the path just randomly, and on average the numbers of ants on every path are the same. Suppose that all ants walk at the same speed and deposit the same amount of pheromone. Since the middle path is the shortest one, ants following this path reach the food point first. Therefore more ants will complete their tour through the middle path in the same period of time, and more pheromone will be deposited in this road correspondingly. When ants return to their nest after they find the food; since there is more pheromone in the middle path, ants will prefer in probability to choose the middle path. This in turn increases the number of ants choosing the middle and shortest path. This is a positive feedback effect with which very soon all ants will follow the shortest path.

## 1.3 Bacterial Foraging Optimization

BFO was introduced by Kevin M. Passino in 2000 for distributed optimization problems [8]. Bacterial Foraging Optimization (BFO) algorithm is a novel evolutionary computation algorithm proposed based on the foraging behavior of Escherichia coli (E. coli) bacteria living in human intestine [15]. The BFO algorithm is a biologically inspired computing technique which is based on mimicking the foraging behavior of E. coli bacteria.

Natural selection tends to remove animals with poor foraging strategies and favors the circulation of genes of those animals that have successful foraging strategies, since they are more likely to enjoy reproductive success. After many generations, poor foraging strategies are either removed or shaped into good ones. This activity of foraging is used in optimization process.

***Framework for BFO algorithm***

- Input the bacterial foraging parameters and independent variable, then specify lower and upper limits of the variables and initiate the elimination-dispersal steps, reproduction and chemotactic.
- Generate the positions of the independent variable randomly for a population of bacteria. Evaluate the objective value of each bacterium.
- By using the tumbling or swimming process, alters the place of the variables for all the bacteria. Perform reproduction and elimination operation.
- If the maximum number of chemotactic, reproduction and elimination-dispersal steps is reached, then output the variable corresponding to the overall best bacterium; Otherwise, repeat the process by modifying the position of the variables for all the bacteria using the tumbling /swimming process .

In this research work, BFO algorithm was hybridized with ACO and a new Hybrid Bacterial Foraging Optimization (HBFO) algorithm was proposed. Both BFO and HBFO algorithm were applied to Admas, Balas and Zawaxk (ABZ), Lawrence (LA) Benchmark problems. The results obtained by HBFO algorithm is compared with BFO, Improved Bacterial Foraging Optimization (IBFO) algorithm and optimal solution of ABZ and LA Benchmark instances.

## 2. RELATED WORKS

Mahanim Omar, Adam Baharum, Yahya Abu Hasan (2006) have proposed a paper about A Job Shop Scheduling Problem (JSSP) using Genetic Algorithm. Job shop problems are widely known as a NP-Hard problem. Here they have generated an primary population arbitrarily including the result obtain by some rules such as shortest processing time and longest processing time. This is used to minimize the objective function [19].

E. Taillard [1989] has proposed a paper about Benchmarks' for Basic Scheduling Problems. In this paper he discussed about 260 scheduling problems whose size is greater than the other examples previously published. Such sizes correspond to real dimensions of industrial problems. In this paper he solved the permutation Flow Shop, Job Shop and Open Shop Scheduling Problems. For Job Shop Scheduling Problem, the machine is allowed any processing order of the jobs. Each job should be processed through the machines in a particular order. The objective of this is minimization of the makespan [10].

Christelle Guéret, Christian Prins [2000], Marc Sevaux have proposed a paper about Applications of optimization with Xpress-MP. In this paper they have discussed about JSSP model formulation, implementation of JSSP and explained JSSP with real time example printing machine. They have used conjunctive and disjunctive constraints between processing operations. Conjunctive represent the precedence between the operations for a single type, and the disjunctive express the fact that a machine can only execute a single operation at a time [7].

Chunguo Wu, Na Zhang, Jingqing Jiang, Jinhui Yang, and Yanchun Liang (2007) described Bacterial Foraging algorithm is a novel evolutionary computation algorithm. This is based on the foraging behaviour of Escherichia Coli bacteria living in human intestine. BFO is essentially a random search algorithm. This enhanced algorithm is applied to Job Shop Scheduling Benchmark problems [8].

Jing Dang, Anthony Brabazon, Michael O'Neill, and David Edition (2008) have proposed a paper about Bacterial Foraging Optimization (BFO) algorithm. This is a biologically inspired computation technique which is based on mimicking the foraging behavior of Escherichia Coli bacteria. During the lifetime of E.coli bacteria, they undergo different stages such as chemotaxis, reproduction and elimination-dispersal. BFO algorithm was implemented various real world problems. Kim suggested that the BFO could be applied to find solutions for difficult engineering design problems [15].

According to S. Subramanian and S. Padma (2011), the selection behaviour of bacteria tends to remove reduced foraging strategies and recover successful foraging strategies. BFO is used to minimizing cost and improves the efficiency simultaneously by using a multi objective based bacterial foraging algorithm [25].

James Montgomery, cardc Fayad and Sarja Petrovic have discussed ACO for Job Shop Scheduling Problems generate solutions by constructing a permutation of the operations, from which a deterministic algorithm can generate the actual schedule. They proposed a paper about Solution Representation for Job Shop Scheduling Problems in ACO. The result produces better solutions [14].

Ashwani Kumar Dhingra has discussed about scheduling problems. He gave a brief explanation about scheduling problems, Significance of Scheduling, Scheduling in a Manufacturing System and Classification of scheduling problems based on requirement generations. Problems up to 200 jobs and 20 machines for instances (DD_SDST_10) developed by Taillard (1993) have been solved and proposed metaheuristics can be tested on other problems [5].

David Applegate, William Cook (1991) have proposed a paper about A Computational Study of the JSSP. They tested performance of JSSP with some sample instances. MT-10 is a well-known 10 by 10 problem of Muth and Thompson; ABZ5 and ABZ6 are two problems from Admas, Balas and Zawaxk; the problems LA19 and LA20 are problems of Lawrence. They compared their results with optimal solution [9].

Jun Zhang, Xiaomin Hu, X.Tan, J.H Zhong and Q. Huang (2006) presented an investigation into the use of an Ant Colony Optimization (ACO) to optimize the JSSP. Ant capable of finding shortest path from their nest to food sources using pheromone trials. Each time ant updates the pheromone trial. The main characteristics of ACO are positive feedback, strength [16].

## 3. HYBRID BACTERIAL FORAGING METHODOLOGY FOR JOB SHOP SCHEDULING PROBLEMS

***The objectives of this research papers are***

- To propose and implement Hybrid Bacterial Foraging Optimization (HBFO) Algorithm to solve JSSP.
- To minimize the makespan of the jobs HBFO is used in scheduling.
- To examine the efficiency of HBFO in solving benchmark instances of JSSP.
- To analyze and compare the performance of the proposed HBFO with BFO in solving JSSP.

### 3.1 Hybrid Bacterial Foraging Optimization(HBFO)

The behavior of ant system is included in tumble part of BFO algorithm, to make it as a HBFO. Each ant builds a tour by repeatedly applying a stochastic greedy rule, which is called the state transition rule.

$$s = \begin{cases} \arg max_{u \cdot J(r)}\{[\tau(r,u)].[\eta(r,u)^\beta]\}, & if\ q \le q_0 (exploitation) \\ S, & otherwise\ (biased\ exploitation) \end{cases} \quad \longrightarrow \quad (1)$$

(r, u) represents an edge between point r and u, and (r, u) stands for the pheromone on edge (r, u). (r, u) is the desirability of edge (r, u), which is usually defined as the inverse of the length of edge (r, u). q is a random number uniformly distributed in [0, 1], $q_0$ is a user-defined parameter with (0 $q_0$ 1), is the parameter controlling the relative importance of the desirability. J (r) is the set of edges available at decision point r. S is a random variable selected according to the probability distribution given below.

$$P(r,s) = \begin{cases} \frac{[\tau(r,s)].[\eta(r,s)^\beta]}{\sum_{u \in J(r)}[\tau(r,u)].[\eta(r,u)^\beta]}, & if\ s\ \epsilon\ J\ (r) \\ 0, & otherwise \end{cases} \quad \longrightarrow \quad (2)$$

The selection strategy used above is also called 'roulette wheel' selection since its mechanism is a simulation of the operation of a roulette wheel [13].

While ant goes for a search it will drop a certain amount of pheromone. It is a continuous process, but we can regard it as a discrete release by some rules. There are two kinds of pheromone update strategies, called local updating rule and the global updating rule.

***Local updating rule***
By applying the local updating rule in an ant tour, it will update the pheromone on the passed edges.

$$\tau(r,s) \quad (1-\rho).\tau(r,s) + \rho.\tau o \quad \longrightarrow \quad (3)$$

Where the coefficient denotes pheromone evaporation, it lies between [0, 1].

***Global updating rule***

Once all ants reached their destination, again it will update the pheromone on the passed edges by applying the global updating rule

$$\tau(r,s) \quad (1-\alpha).\tau(r,s) + \alpha. \ .\tau(r,s) \quad \longrightarrow \quad (4)$$

Where

$$.\tau(r,s) \begin{cases} (L_{gb})^{-1}, & if\ (r,s)\ :global - best - tour \\ 0, & otherwise \end{cases} \quad \longrightarrow (5)$$

Here 0< <1 is the pheromone decay parameter, and $L_{gb}$ is the length of the globally best tour from the beginning of the trial. (r; s) is the pheromone addition on edge (r, s). We can see that only the ant that finds the global best tour can achieve the pheromone increase [13].

In BFO, the objective is to find the minimum of J( ), $R^D$, where we do not have the gradient information J( ). Suppose is the position of the bacterium and J( ) represents a nutrient profile, i.e.,J( ) < 0, J( )=0and J( ) > 0 represent the presence of nutrients, a neutral medium and noxious substances respectively. The bacterium will try to move towards growing concentrations of foods (i.e. find lower values of J), search for ways out of neutral media and avoid injurious substances (away from positions where J > 0). It implements a type of biased random walk.

The mathematical swarming (cell-cell signalling) function can be represented by:

$$J_{cc}\left(\theta^i, \theta\right) =$$
$$\begin{cases} -M\left(\sum_{k=1}^{s} e^{-w_a||\theta^i-\theta^k||^2} - \sum_{k=1}^{s} e^{-w_r||\theta^i-\theta^k||^2}\right) With\ swarming \\ 0 \qquad\qquad\qquad\qquad\qquad\qquad\qquad\qquad\qquad No\ swarming \end{cases} \rightarrow (6)$$

Where $\|.\|$ represents the Euclidean norm, $W_a$ denotes the width of the attractant and $W_r$ denotes the width of the repellent signals, M is the magnitude of the cell-cell signaling effect [12].

The above State Transition rule of ant in ACO is included in the tumble. HBFO methodology is implemented with no swarming effect (ie) $j_{cc}$=0 [15]. Here time is considered as cost. During the lifetime of E-Coli bacteria they undergo different stages such as Chemotactics, Reproduction and Elimination-Dispersal.When compared with ACO and BFO, HBFO achieves high level of SHA1PRNG algorithm incase of reproduction, elimination-dispersal.

### *HBFO Algorithm*

**for** *Elimination-dispersal loop* **do**
    **for** *Reproduction loop* **do**
        **for** *Chemotaxis loop* **do**
            **for** *Bacterium i* **do**
                Tumble: Generate a secure random vector q  decimal value.
                **If** q < $q_0$ **then**
                    Generate a secure random vector l  operation,
                    according to pheromone value ph[job][operation] based
                    on equation 1.
                **Else**
                    Generate a secure random vector l  operation,
                    according to pheromone value ph[job][operation] based
                    on equation 2.
                **end**
                **Move**: Generate a secure random vector $l_{new}$  operation.
                    **Swim**:
                    **if** time[job][l] < time[job][ $l_{new}$] **then**
                      current_operation = 1
                    **Else**
                      current_operation = $1_{new}$
                    **end**
            **end**
        **end**
        Sort bacteria in order of ascending time $J_{st}$. The $S_r = S/2$ bacteria with The highest
        *J* value die and other $S_r$ bacteria with the best value split Update value of *J* and $J_{st}$
        accordingly.
    **end**
    Eliminate and disperse the bacteria to random locations on the optimization domain with
    probability $p_{ed}$. Update corresponding *J* and $J_{st}$.
**End**

*The parameters are described below in the Table I*

**Table 1: Description of Parameters**

| Parameter Name | Description |
|---|---|
| $J_{cc}$ | Health of bacterium i |
| $J^i_{health}$ | Health of bacterium i |
| L | Counter for elimination- dispersal step |
| $P_{ed}$ | Probability of occurrence of elimination-dispersal events |
| S | Population of the E. coli bacteria |
| attract | Width of attractant |
| repellant | Width of repellent |

# 4. IMPLEMENTATION RESULTS AND DISCUSSION

This paper discusses and compares the result of the implementation of BFO and proposed HBFO algorithm in solving the Benchmark instances of FSSP

Admas, Balas and Zawaxk (ABZ), Lawrence (LA) Benchmark problems [30] for JSSP were solved in this research work. JSSP Benchmark instances were taken from Operations Research (OR) Library to test the efficiency of proposed MBFO. The proposed HBFO algorithm gave feasible solution for most runs for the constant values **=0.1, =1.0, =0.1, q0=0.8, =0.5.** The result obtained by proposed HBFO algorithm was compared with BFO and IBFO algorithm. The HBFO algorithm gave a minimum makespan, when compared with the makespan obtained by BFO and IBFO. The computational results are given below in tables

## 4.1 Results for 10*5 and 15*5 Lawrence Instances

LA Benchmark problems for JSSP were solved in this research work. The optimal solution obtained from proposed HBFO algorithm and BFO algorithm were compared with Improved Bacterial Foraging Optimization (IBFO) value. The results are shown in Table II. The figure II shows the graphical representation of Table II.

**Table 2: Results for 10* 5 and 15* 5 Lawrence Instances**

| INSTANCE | SIZE | BFO | HBFO | IBFO[8] |
|---|---|---|---|---|
| LA 01 | 10 * 5 | 694 | **693** | 666 |
| LA 02 | 10 * 5 | 692 | **683** | 668 |
| LA 03 | 10 * 5 | 639 | **634** | 617 |
| LA 04 | 10 * 5 | 641 | **624** | 604 |
| LA 05 | 10 * 5 | 593 | **593** | 593 |
| LA 06 | 15 * 5 | 926 | **926** | 926 |
| LA 07 | 15 * 5 | 923 | **903** | 890 |
| LA 08 | 15 * 5 | 877 | **873** | 863 |
| LA 09 | 15 * 5 | 954 | **951** | 951 |
| LA 10 | 15 * 5 | 958 | **958** | 958 |

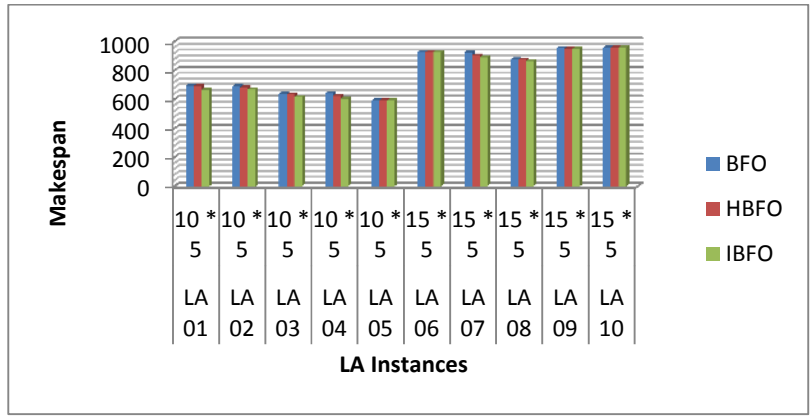

Figure 2: Graphical representation of results for 10* 5 and 15* 5 Lawrence Instances

## 4.2 Results for 20*15 and 20*10 Lawrence Instances

The optimal solution obtained from proposed HBFO algorithm and BFO algorithm were compared with optimal value of ABZ and LA Instances are shown in Table III. The figure III shows the graphical representation of Table III.

**Table 3: Results for 20*15 and 20*10 Lawrence Instances**

| INSTANCE | SIZE | BFO | HBFO | OPTIMAL[9] |
|----------|------|-----|------|------------|
| ABZ 7 | 20 * 15 | 787 | **784** | 668 |
| ABZ 8 | 20 * 15 | 822 | **792** | 687 |
| ABZ 9 | 20 * 15 | 856 | **840** | 707 |
| LA 27 | 20 * 10 | 1455 | **1446** | 1269 |
| LA 29 | 20 * 10 | 1409 | **1390** | 1195 |

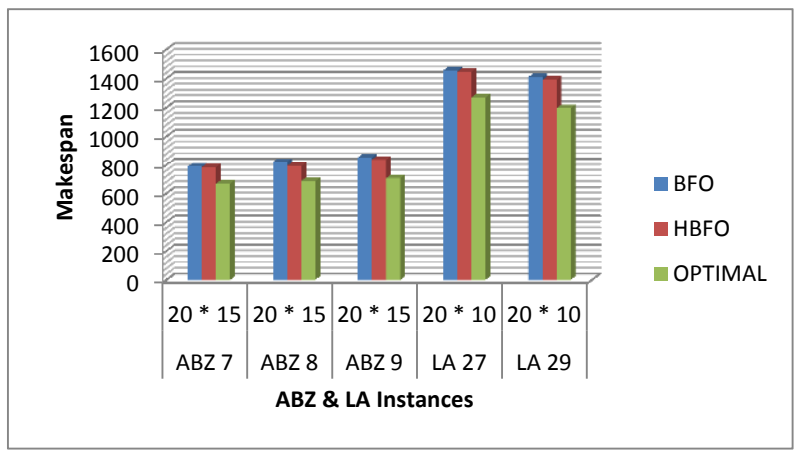

Figure 3: Graphical representation of results for 20*15 and 20*10 Lawrence Instances

**4.3 Results for 10*10 ABZ and 15*10 LA Instances**

BFO algorithm and Optimal values are compared with proposed HBFO algorithm of ABZ 10*10 and LA instances 15*10 are shown in Table IV. The figure IV shows the graphical representation of Table IV.

**Table 4: Results for 10*10 ABZ and 15*10 LA Instances**

| INSTANCE | SIZE | BFO | HBFO | OPTIMAL[9] |
|----------|------|-----|------|------------|
| ABZ 5 | 10 *10 | 1323 | **1321** | 1234 |
| ABZ 6 | 10 *10 | 1012 | **979** | 943 |
| LA 19 | 10 *10 | 926 | **894** | 842 |
| LA 21 | 15 *10 | 1247 | **1207** | 1053 |
| LA 24 | 15 *10 | 1102 | **1102** | 935 |
| LA 25 | 15 *10 | 1147 | **1131** | 977 |

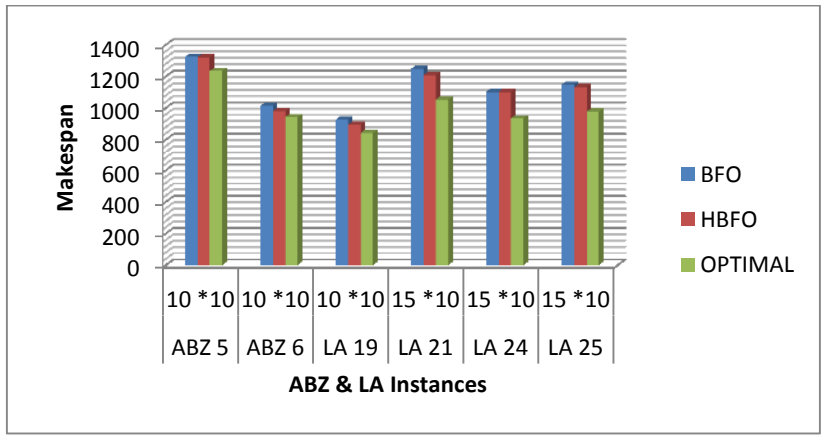

Figure 4: Graphical representation of results for 10*10 ABZ and 15*10 LA Instances

# 5.  CONCLUSIONS

It was clearly understandable that proposed HBFO algorithm gave the best makespan for ABZ, LA instances when compared with BFO algorithm and. From the implementation of this research work, it was observed that the proposed HBFO was effective than BFO algorithm in solving JSSP. The proposed HBFO algorithm can also be used for higher instances of size and this HBFO algorithm will surely be able to achieve the best makespan for more number of iterations.

The ability of the proposed HBFO algorithm was investigated through the performance of several runs on well-known test problems of different sizes, which were taken from OR library, which is the primary repository for such problems. The results obtained by the proposed HBFO for JSSP are much better and highly comparable to the results obtained by BFO and IBFO algorithms. The proposed HBFO for JSSP can be improved to achieve optimal solution by including the swarming technique and also by modifying constant values used in the algorithms. As a future work Flexible Job Shop Scheduling problems can also be solved using proposed HBFO algorithm.

## Authors

**Mr. S. Narendhar** received his B. Sc Degree in Computer Technology from Anna University, Coimbatore, India in the year 2009 and Masters Degree (MCA) in Computer Applications from Bharathiar University, Coimbatore, India in the year 2012. His area of interest includes Agent based computing and Bio-inspired computing. He has attended National Conferences.

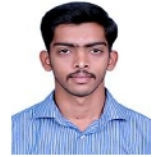

**Mrs. T. Amudha** received her B.Sc Degree in Physics, Masters Degree (MCA) in Computer Applications and M.Phil in Computer Science in 1995, 1999, and 2003 respectively, from Bharathidasan University, India. She has qualified UGC-NET for Lectureship in 2003 and is currently pursuing her doctoral research at Bharathiar University in the area of Agent Systems. She is currently working as Asst. Professor in the Department of Computer Applications, Bharathiar University, Coimbatore and has 13 years of academic experience. She has more than 20 research publications for her credit in International/ National Journals & Conferences. Her area of interest includes Software Agents, Bio-inspired computing and Grid computing.

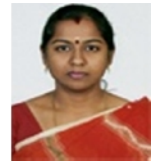